\documentclass{article}
\usepackage{spconf,amsmath,graphicx}
\usepackage{hyperref} 
\usepackage{enumitem}
\setlist{nosep, leftmargin=14pt}
\usepackage{amsmath}
\usepackage{amssymb}
\usepackage{mathtools}
\usepackage{amsthm}
\usepackage[table]{xcolor}
\definecolor{vlmgray}{RGB}{226, 226, 232}
\theoremstyle{plain}

\theoremstyle{definition}

\theoremstyle{remark}

\usepackage{graphicx}
\usepackage{url}
\usepackage{array}
\usepackage{multirow}
\usepackage{booktabs} 
\usepackage{wrapfig}
\usepackage{hyperref}
\usepackage{url}
\newcolumntype{P}[1]{>{\centering\arraybackslash}p{#1}} 
\newcolumntype{R}[1]{>{\raggedleft\arraybackslash}p{#1}} 
\newcolumntype{L}[1]{>{\raggedright\arraybackslash}p{#1}} 

\newcommand{\msinline}[2]{%
  #1\raisebox{-0.5ex}{\tiny$\pm\,#2$}%
}

\definecolor{vlmgray}{HTML}{F0F0F0}
\definecolor{resultblue}{HTML}{0072B2}
\definecolor{changeRed}{HTML}{C44E52}
\definecolor{changeGreen}{HTML}{25845B}
\definecolor{ablationgreen}{HTML}{b2f2bb}

\newlength\savewidth

\definecolor{lower}{RGB}{58,71,172}     
\definecolor{higher}{RGB}{166,67,58}    
\definecolor{myyellow}{RGB}{244,164,96}
\definecolor{myblue}{RGB}{65,105,225}   
\definecolor{mygreen}{RGB}{47,139,87}   
\definecolor{Highlight}{HTML}{39b54a}  
\definecolor{lightblue}{RGB}{51,153,204}

\usepackage{mwe} 

\title{STRATA: Reinforcement-driven Active Sampling with Tri-paradigm Adaptation for Efficient 3D Spatial Transcriptomics}

\name{
\begin{tabular}{c}
Junchao Zhu$^{1}$\qquad
Ruining Deng$^{2}$\qquad
Chongyu Qu$^{1}$\qquad 
Zhengyi Lu$^{1}$\qquad 
Yanfan Zhu$^{1}$\qquad \\
Yuechen Yang$^{1}$\qquad 
Yu Wang$^{3}$\qquad
Shilin Zhao$^{3}$\qquad 
Haichun Yang$^{3}$\qquad
Yuankai Huo$^{1,{\star}}$\thanks{Email: \href{email:yuankai.huo@vanderbilt.edu}{yuankai.huo@vanderbilt.edu}}
\end{tabular}
}

\address{
$^{1}$ Vanderbilt University, Tennessee, USA \\
$^{2}$ Weill Cornell Medicine, New York, USA \\
$^{3}$ Vanderbilt University Medical Center, Tennessee, USA
}
\begin{document}
%
\maketitle
\begin{abstract}
Spatial transcriptomics (ST) enables joint characterization of tissue morphology and gene expression. However, 3D volumetric ST acquisition remains expensive. AI-driven methods offer a low-cost alternative but remain limited in accuracy, while recent 2D section-wise ST acquisition still rely on dense grid sampling, resulting in redundant measurements and undersampling of biologically informative regions. To bridge the gap, we propose STRATA, a reinforcement-driven active sampling framework with tri-paradigm adaptation for reconstructing dense 3D transcriptomic maps from sparse voxel-wise ST acquisition. STRATA constructs a 3D graph integrating H\&E morphology with single-cell priors and employs a reinforcement learning agent to estimate morphological sparsity and transcriptomic informativeness, enabling budget-aware selection of informative spots. Given these sparse measurements, an adaptive tri-paradigm predictor combines retrieval, regression, and cross-layer imputation through lightweight fusion, dynamically weighting their contributions. Experiments across datasets demonstrate improved sampling efficiency and accuracy, particularly under low-budget settings. Code is available at \href{https://github.com/Junchao-Zhu/STRATA}{STRATA}.
\end{abstract}

\section{Introduction}
Spatial transcriptomics (ST) enables in situ profiling of tissue morphology and gene expression~\cite{he2020integrating,staahl2016visualization}. However, ST remains expensive, with sequencing a single section costing thousands of dollars, making large-scale 3D tissue modeling financially impractical. Although current methods can infer gene expression directly from histology, their performance still depends heavily on sufficient ST supervision. Yet conventional ST acquisition relies on fixed-grid sampling~\cite{staahl2016visualization,andersson2021spatial}, despite strong spatial continuity in H\&E morphology and gene expression~\cite{zeng2022spatial,zhu2025asign}, resulting in redundant measurements and inefficient use of limited sequencing budgets. This redundancy becomes more pronounced in 3D tissues, where biologically informative patterns are often concentrated in limited regions, motivating adaptive selection of spots with high morphological and transcriptomic informativeness. Meanwhile, existing predictors are predominantly 2D and single-paradigm~\cite{chen2025visual,xie2024spatially}, overlooking cross-layer morphology--expression continuity and inheriting limitations of regression or retrieval alone. These motivate a unified framework that jointly optimizes 3D ST acquisition and adaptive prediction by exploiting cross-layer continuity and complementary inference paradigms.

\begin{figure*}[tbhp]
    \centering
    \includegraphics[width=0.85\linewidth]{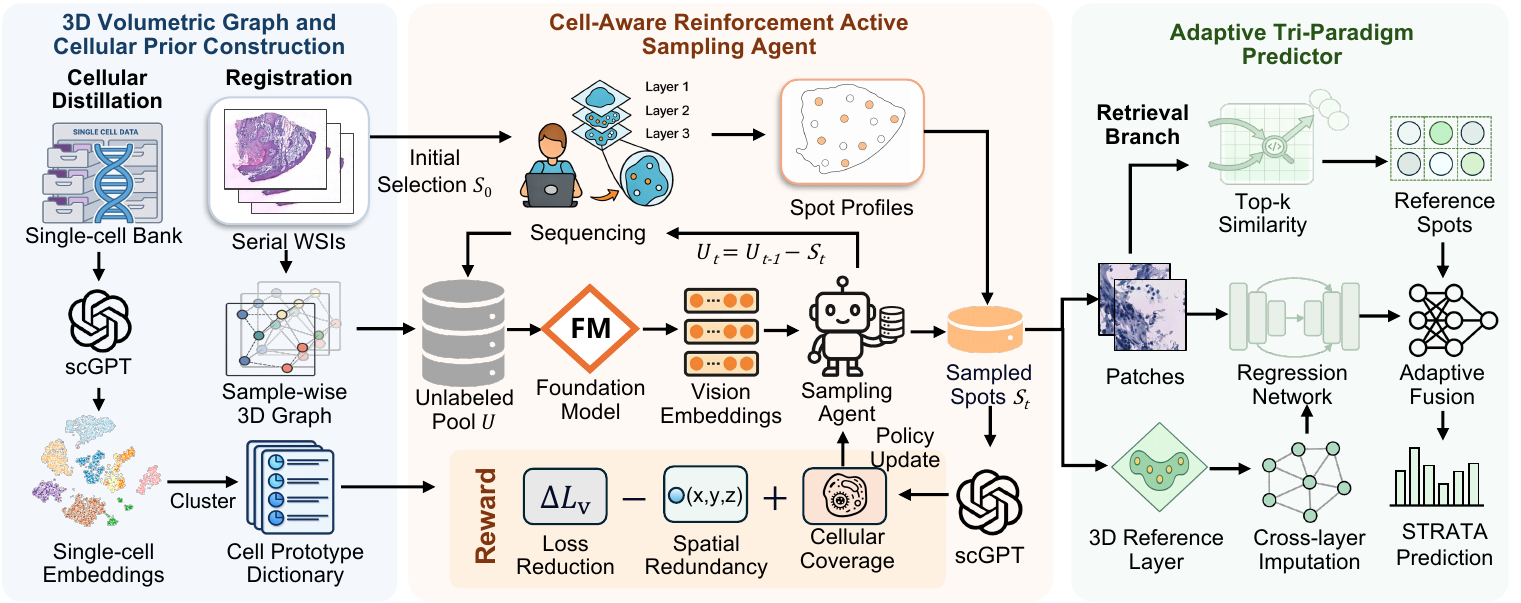}
    \caption{\textbf{Overview of STRATA.}  We first align serial H\&E sections into a 3D voxel space and construct a cellular prototype dictionary from single-cell reference to introduce biological priors. We then develop a cell-aware RL sampling agent that evaluates candidate spots by morphological sparsity and bio-informativeness, and optimizes selection with validation loss reduction, 3D redundancy penalty, and prototype coverage bonus. Finally, an adaptive tri-paradigm predictor integrates visual retrieval, regression, and 3D cross-layer imputation via a lightweight head to generate high-fidelity expression predictions.}
    \label{fig:framework}
\end{figure*}
To bridge these gaps, we propose STRATA, a unified reinforcement-driven active sampling framework with tri-paradigm adaptation that reformulates ST modeling as a joint sampling-and-prediction problem over 3D tissue volumes under constrained sequencing budgets. STRATA constructs a volumetric graph through histological alignment and incorporates single-cell references as biological priors to guide spatial modeling. Based on this representation, a reinforcement learning agent adaptively selects informative spots according to morphological sparsity and transcriptomic informativeness, prioritizing regions with higher biological value, and iteratively updates its sampling policy as new measurements become available. Meanwhile, a tri-paradigm predictor integrates retrieval, regression, and 3D cross-layer imputation through lightweight soft routing to exploit their complementary inductive biases. Our contributions are threefold:
\begin{itemize}
\item We propose STRATA that reformulates isolated 2D ST acquisition and prediction as budget-aware 3D co-optimization, enabling dense sparse voxel-wise measurements under limited sequencing budgets.
\item We develop a reinforcement-driven active sampling strategy on a 3D voxel graph integrating H\&E morphology and single-cell priors, enabling iterative selection of spots with high morphological sparsity and informativeness.
\item We introduce an adaptive tri-paradigm predictor that soft-routes across retrieval, regression, and 3D cross-layer imputation to improve prediction under limited budgets.
\end{itemize}

\section{Method}
\subsection{3D Graph and Cellular Prior Construction}
\noindent\textbf{3D graph construction.}
We first register sections into a 3D system using the middle as the reference with affine registration and refinement. For each spot \(i\), a pretrained vision encoder encodes its H\&E patch into morphological \(f_i\). All spots form a 3D graph \(\mathcal{G}\). Within each section, each spot is connected to its \(k_{\mathrm{intra}}\) nearest neighbors based on Euclidean distance. For cross-section connections, we identify overlapping candidates and rank them by \(\cos(f_i,f_j)\). If fewer than \(k_{\mathrm{cross}}\) candidates, additional non-overlapping neighbors are selected using \((1-\alpha)\cos(f_i,f_j)+\alpha(1-d_{ij}/d_{\max})\), where \(d_{ij}\) denotes spatial distance, \(d_{\max}\) is for normalization, and \(\alpha\) balances similarity and spatial proximity. Cross-section edge weights are then defined as \(w^{\mathrm{cross}}_{ij}=\cos(f_i,f_j)\), reducing effect of registration errors.

\noindent\textbf{Cellular state prototypes.}
Using a large-scale single-cell reference, pretrained foundation model embeds reference cell $j$ and sequenced spot $i$ into a shared space as $q_j$ and $z_i$, respectively. MiniBatchKMeans clusters $N$ reference embeddings into $C$ prototypes by solving $\min_{\{\mu_c\}_{c=1}^{C}}\sum_{j=1}^{N}\min_c\|q_j-\mu_c\|_2^2$, where $\mu_c$ denotes a prototype centroid. Let $\mathcal{C}_c$ denote the reference cells assigned to prototype $c$. The resulting dictionary $\mathcal{D}=\{(\mu_c,\pi_c)\}_{c=1}^{C}$ records prototype prevalence $\pi_c=|\mathcal{C}_c|/N$, thereby encouraging the policy to explore biologically underrepresented regions.

\subsection{Cell-Aware Active Sampling Agent on 3D Graph}
\noindent\textbf{Sampling policy with cellular priors and trajectory memory.}
We formulate sampling on $\mathcal{G}$ as a finite-horizon decision process with state $s_t=(\mathcal{G},m_t,\mathcal{S}_t,b_t)$, where $\mathcal{U}$ is the full candidate set, $m_t\in\{0,1\}^{|\mathcal{U}|}$ marks unsampled spots with $1$, $\mathcal{S}_t\subset\mathcal{U}$ is the set selected by step $t$, and $b_t=(B-|\mathcal{S}_t|)/B$ is the remaining budget ratio. The action $a_t\in\mathcal{U}\setminus\mathcal{S}_t$ is drawn from a masked softmax policy.  Static features $\phi_i^{\text{frz}}$ depend only on H\&E and graph geometry, including $f_i$ and mean within- and cross-section neighbor embeddings $\bar f_i^{\text{intra}}$ and $\bar f_i^{\text{cross}}$. Dynamic features $\phi_i^{\text{dyn}}(t)$ incorporate cellular soft assignment $c_i$, $b_t$, and mean cross-section similarity $\overline{w}^{\text{cross}}_{i,\mathcal{S}_t}$ to selected spots; $z_i$ is included only after sequencing. 

The policy $\pi_\theta$ scores unsampled candidates. A lightweight cellular state prediction head $g_\phi:f_i\mapsto\Delta^{C-1}$ predicts prototype soft assignments $c_i$, where $\Delta^{C-1}$ is the probability simplex. After sequencing, the nearest prototype to $z_i$ provides a hard label for cross-entropy updates on accumulated samples, progressively distilling single-cell knowledge. A frozen projection produces $h_i^{\text{static}}$, keeping reinforcement learning gradients focused on the scoring head. Trajectory context $h_t^{\text{ctx}}$ averages selected static embeddings. The policy combines these representations as: 
\begin{equation}
\begin{gathered}
\ell_{t,i}=\mathrm{MLP}_\theta\!\big([h_i^{\text{static}}\Vert c_i\Vert\phi_i^{\text{dyn}}(t)\Vert h_t^{\text{ctx}}]\big),\\
\pi_\theta(a_t=i\mid s_t)=\frac{m_{t,i}\exp(\ell_{t,i})}{\sum_jm_{t,j}\exp(\ell_{t,j})}.
\end{gathered}
\end{equation}
$\Vert$ denotes concatenation and $\theta$ denotes policy parameters. The mask excludes previously selected spots, while trajectory context and cross-section similarity reduce redundancy.

\noindent\textbf{Multi-objective reward and policy optimization.}
We train $\pi_\theta$ by reinforcement learning with a per-step reward combining validation-loss reduction, a 3D redundancy penalty, and a coverage bonus for underrepresented prototypes:
\begin{equation}
r_t=R^{\text{round}}-\gamma\,\overline{w}^{\text{cross}}_{a_t,\mathcal{S}_t}
+\eta_{\text{sc}}\log\!\frac{B}{1+n^{\text{sc}}_{k^\star(a_t)}}.
\end{equation}
After sequencing the chosen spot, its nearest prototype is $k^\star(a_t)=\arg\min_c\|z_{a_t}-\mu_c\|_2$. Here, $n^{\text{sc}}_k=|\{i\in\mathcal{S}_t:k^\star(i)=k\}|$ counts spots assigned to prototype $k$ before the current selection; $\gamma$ and $\eta_{\text{sc}}$ weight redundancy and coverage, respectively. Sampling proceeds over $R$ rounds. Each round commits a batch and updates the predictor from $\theta^{-}$ to $\theta^{+}$ on $\mathcal{S}_t$. On a deterministically hashed validation subset $\mathcal{V}^{\text{ps}}\subset\mathcal{S}$, the primary reward is $R^{\text{round}}=\mathcal{L}_{\mathcal{V}^{\text{ps}}}(\theta^{-})-\mathcal{L}_{\mathcal{V}^{\text{ps}}}(\theta^{+})$, so a reduction in validation loss yields a positive reward.
All actions in the round share this reward. Exponential-moving-average normalization of $r_t$ yields advantage $A_t$. To alleviate gradient sparsity, we replay actions under updated policy parameters, recomputing log-probabilities to accumulate gradients. The policy objective, where $\mathcal{H}$ is entropy and $\beta$ decreases during training to transition from exploration to convergence, as:
\begin{equation}
\mathcal{L}_\pi=-\mathbb{E}_t\!\left[
A_t\log\pi_\theta(a_t\mid s_t)
+\beta\,\mathcal{H}\!\big(\pi_\theta(\cdot\mid s_t)\big)
\right],
\end{equation}
\subsection{Adaptive Tri-Paradigm Predictor}
\noindent\textbf{Multi-paradigm modeling and optimization.}
Three parallel branches provide complementary predictions. Retrieval uses $\mathcal{S}$ as a global reference pool and encodes patches and expression into a shared semantic space. The top-$K$ cosine-matched references are softmax-weighted to obtain $\hat y_i^{\text{ret}}$. Regression uses pretrained CONCH and a lightweight head to predict $\hat y_i^{\text{reg}}$, providing predictions for regions poorly covered by the reference pool. The 3D imputation branch incorporates cross-section structure through $\mathcal{G}$. It aggregates sequenced neighbors $\mathcal{N}_i^{\text{cross}}\subset\mathcal{S}$ with normalized attention weights $\alpha_{ij}\propto w^{\text{cross}}_{ij}\exp(\cos(f_i,f_j))$, yielding $\hat y_i^{\text{3d}}=\sum_j\alpha_{ij}y_j$ as a sample-level consistency prior, where $y_j$ is measured expression.

Retrieval aligns paired visual and expression embeddings through a symmetric contrastive loss. Let $v_i$ and $h_i$ denote paired visual and expression embeddings in a batch of $N$ spots, with contrastive temperature $\tau=0.07$. Each branch computes its loss independently with decoupled gradients:
\begin{equation}
\begin{aligned}
\mathcal{L}_{Ret} \;&=\; -\frac{1}{2N}\sum_{i=1}^{N}\!\bigg[\log\frac{\exp(v_i^\top h_i / \tau)}{\sum_{j=1}^{N}\exp(v_i^\top h_j / \tau)} + (v\leftrightarrow h)\bigg], \\
\mathcal{L}_{Reg} \;&=\; \mathrm{MSE}\!\big ( \hat y^{reg}_i,\, y_i\big) \;+\; \lambda_{\text{con}}\,\mathrm{MSE}\!\big(\hat y^{reg}_i,\, \hat y^{\text{3d}}_i).
\end{aligned}
\end{equation}
Here, $(v\leftrightarrow h)$ denotes the second logarithmic term obtained by exchanging $v$ and $h$. MSE denotes mean squared error. The regression loss includes a 3D consistency term weighted by $\lambda_{\text{con}}$, as a soft teacher to transfer structural supervision.

\noindent\textbf{Adaptive expert routing fusion.}
We introduce a lightweight adaptive fusion head to route each spot toward an optimal combination of the three branches. After training, we cache their predictions and 3D confidence on the sampled set $\mathcal{S}$ and train the fusion head on these outputs. Two global learnable scalars $w_3,w_r\in\mathbb{R}$, together with spot-level confidence $\rho_i=\max_j\alpha_{ij}$, determine the branch weights. 

Routing proceeds in two stages. First, the 3D contribution is $\alpha_i^{\text{3d}}=\rho_i\sigma(w_3)$, where $\sigma$ denotes sigmoid. The confidence $\rho_i$ adaptively scales this contribution, reducing the influence of 3D imputation when confidence is low. The remaining weight $1-\alpha_i^{\text{3d}}$ is split as $\alpha_i^{\text{ret}}=(1-\alpha_i^{\text{3d}})\sigma(w_r)$ and $\alpha_i^{\text{reg}}=(1-\alpha_i^{\text{3d}})(1-\sigma(w_r))$, using $\sigma(w_r)$ to balance retrieval and regression. These weights adapt the three-branch combination for each spot, yielding the final prediction:
\begin{equation}
\hat y_i
=\alpha_i^{\text{ret}}\hat y_i^{\text{ret}}
+\alpha_i^{\text{reg}}\hat y_i^{\text{reg}}
+\alpha_i^{\text{3d}}\hat y_i^{\text{3d}}.
\end{equation}
\section{Experiment and Result}
\noindent\textbf{Experimental Setup.} We evaluate on public HER2~\cite{andersson2021spatial}, Breast~\cite{he2020integrating}, cSCC~\cite{ji2020multimodal}, and private Kidney cohorts using spot-centered $224\times224$ H\&E patches and $\log(1+x)$-transformed counts of the top 300 highly variable genes, following BLEEP~\cite{xie2024spatially}. Prototype distillation uses $\sim$2M Kidney cells, $\sim$3M cells for HER2/Breast, and cSCC single-cell references. Prediction baselines are ST-Net~\cite{he2020integrating}, EGN~\cite{yang2023exemplar}, His2ST~\cite{zeng2022spatial}, BLEEP~\cite{xie2024spatially}, mclSTExp~\cite{min2024multimodal}, OmiCLIP~\cite{chen2025visual}, and ASIGN~\cite{zhu2025asign}. Sampling baselines share a predictor and include Random, Learning Loss (L.Loss)~\cite{yoo2019learning}, MaxHerding (Maxh)~\cite{bae2024generalized}, ProbCover~\cite{yehuda2022active}, and spatial Poisson sampling. We report Pearson correlation coefficient (PCC) and mean squared error (MSE). Experiments use one RTX A6000 and patient-level cross-validation, using each patient's middle slice for training and 3D imputation anchoring and remaining slices for evaluation~\cite{zhu2025asign}. Learning rates are $5\times10^{-4}$ (retrieval/fusion), $5\times10^{-5}$ (regression), and $3\times10^{-4}$ (policy). Retrieval/regression use weight decay $10^{-4}$ and batch size 64; retrieval uses top-$k{=}25$, fusion trains for 30 epochs, and policy hidden dimension is 128. Sampling runs every 5 epochs after 5 warmup epochs through epoch 50, with $\gamma=0.05$, $\eta_{\text{sc}}=0.03$, $\kappa=1.3$, and $\beta$ annealed from 0.1 to 0.01. All methods share splits and protocols at budget ratios $\{0.1,0.3,0.5,0.75\}$.
\begin{table}[tbhp]
\centering
\scriptsize
\caption{\textbf{Quantitative comparison on gene expression prediction.} STRATA achieves the best overall performance.}
\label{tab:gene_prediction}
\renewcommand{\arraystretch}{1.0}
\setlength{\tabcolsep}{2.5pt}
\resizebox{\columnwidth}{!}{%
\begin{tabular}{l|cc@{\hspace{3pt}}cc@{\hspace{3pt}}cc@{\hspace{3pt}}cc}
\toprule
\multirow{2}{*}{\textbf{Method}}
& \multicolumn{2}{c}{\textbf{HER2~\cite{andersson2021spatial}}}
& \multicolumn{2}{c}{\textbf{Breast~\cite{he2020integrating}}}
& \multicolumn{2}{c}{\textbf{cSCC~\cite{ji2020multimodal}}}
& \multicolumn{2}{c}{\textbf{Kidney}} \\
\cmidrule(lr){2-3}
\cmidrule(lr){4-5}
\cmidrule(lr){6-7}
\cmidrule(lr){8-9}
& MSE$\downarrow$ & PCC$\uparrow$
& MSE$\downarrow$ & PCC$\uparrow$
& MSE$\downarrow$ & PCC$\uparrow$
& MSE$\downarrow$ & PCC$\uparrow$ \\
\midrule

\rowcolor{vlmgray}
\multicolumn{9}{c}{\textbf{\textit{2D Monolithic Models}}} \\

ST-Net~\cite{he2020integrating}
& 0.527 & 0.271
& 0.623 & 0.152
& 1.515 & 0.137
& 0.728 & 0.227 \\

His2ST~\cite{zeng2022spatial}     
& 0.603  & 0.244 
& 0.740  & 0.148 
& 1.667 & 0.117 
& 0.776 & 0.148 \\

EGN~\cite{yang2023exemplar}
& 0.603 & 0.268
& 0.677 & 0.124
& 1.601 & 0.137
& 0.770 & 0.118 \\

mclSTExp~\cite{min2024multimodal}
& 0.593 & 0.259
& 0.745 & 0.128
& 1.804 & 0.142
& 0.871 & 0.201 \\

OmiCLIP~\cite{chen2025visual}
& 0.469 & 0.305
& 0.662 & 0.140
& 1.487 & 0.165
& 0.732 & 0.282 \\

BLEEP~\cite{xie2024spatially}
& 0.456 & 0.287
& 0.590 & 0.169
& 1.445 & 0.186
& 0.682 & 0.325 \\

\rowcolor{vlmgray}
\multicolumn{9}{c}{\textbf{\textit{3D Volumetric Models}}} \\

ASIGN~\cite{zhu2025asign}
& 0.472 & 0.300
& 0.537 & 0.208
& 1.518 & 0.197
& 0.876 & 0.244 \\

\underline{\textbf{STRATA (Ours)}}
& \cellcolor{resultblue!10}\textbf{0.356}
& \cellcolor{resultblue!10}\textbf{0.361}
& \cellcolor{resultblue!10}\textbf{0.471}
& \cellcolor{resultblue!10}\textbf{0.230}
& \cellcolor{resultblue!10}\textbf{1.337}
& \cellcolor{resultblue!10}\textbf{0.227}
& \cellcolor{resultblue!10}\textbf{0.660}
& \cellcolor{resultblue!10}\textbf{0.361} \\

\bottomrule
\end{tabular}%
}
\end{table}

\noindent\textbf{Gene Expression Prediction.} We systematically evaluate STRATA against SOTA methods across cohorts. As shown in Table~\ref{tab:gene_prediction}, STRATA achieves the highest PCC and lowest MSE across all cohorts. On HER2, compared with the strongest baseline, PCC increases from 0.314 to 0.361 (15.0\%), while MSE decreases from 0.434 to 0.356 (18.0\%). By operating on a 3D voxel representation and adaptively fusing three complementary modeling paradigms, STRATA enables more effective integration of cross-paradigm information, leading to improved overall prediction performance.

\begin{figure}[thbp]
    \centering
\includegraphics[width=\columnwidth]{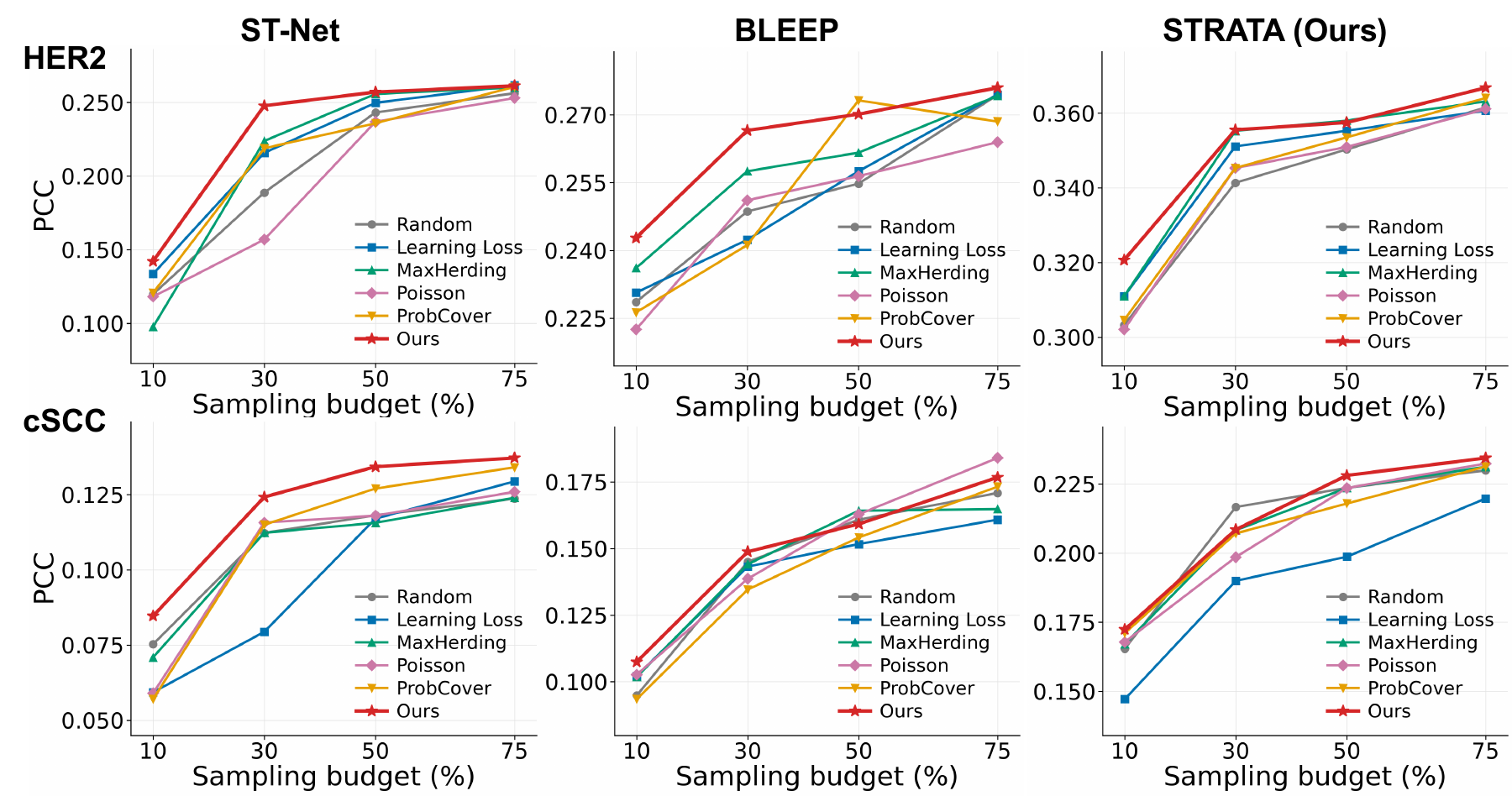}
\caption{\textbf{Comparison of active sampling strategies}, reporting PCC across backbones and datasets. STRATA achieves superior or competitive performance in most settings, with notable gains in the low-budget regime.}
    \label{fig:active_sampling}
\end{figure}

\noindent\textbf{Active Sampling under Limited Budgets.} Figure~\ref{fig:active_sampling} compares sampling strategies across three predictors on HER2 and cSCC. Our policy achieves the highest PCC at a 10\% budget. With ST-Net on cSCC, it improves PCC from 0.075 to 0.085 (12.4\%), demonstrating comparable accuracy with fewer measurements. Gains extend to BLEEP and STRATA across both evaluated cohorts. This stability is attributed to the complementary modeling between the 3D graph and the tri-branch predictor, enabling STRATA to maintain reliable performance in clinically realistic low-budget scenarios with limited measurement resources for practical deployment.

\begin{figure}[tbhp]
    \centering
    \includegraphics[width=0.9\columnwidth]{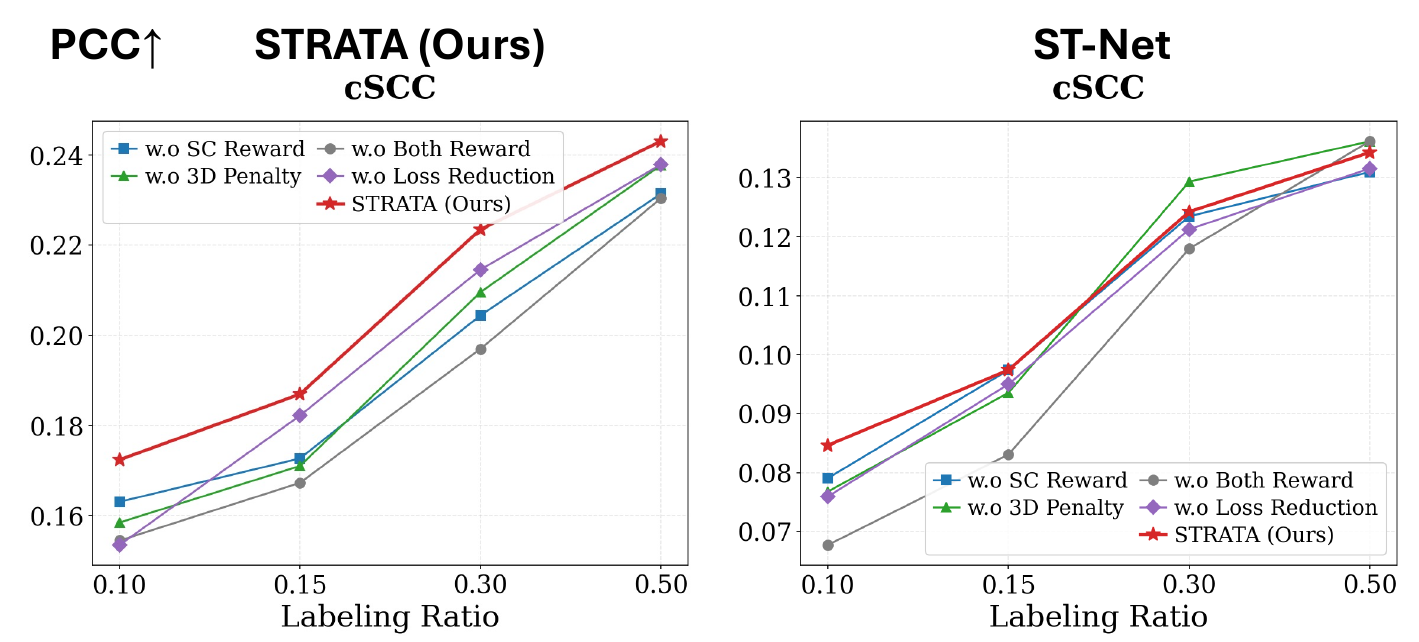}
    \caption{\textbf{Reward ablation on the RL sampling agent.} Each curve denotes one reward variant.}
    \label{fig:abl_reward}
\end{figure}

\noindent\textbf{Predictor Ablation.} Table~\ref{tab:ablation_branches} shows that branch effectiveness varies across cohorts. Regression achieves the highest single-branch PCC on HER2 (0.328), whereas 3D imputation performs best on cSCC (0.200), motivating their complementary integration. Adding 3D consistency supervision improves PCC from 0.314 to 0.328 on HER2. Finally, the full tri-branch STRATA achieves the best performance, indicating that the branches are complementary and a lightweight fusion head with learnable global preferences enables effective integration across heterogeneous cohorts.

\noindent\textbf{Sampling Reward Ablation.} We ablate the RL agent’s reward with four variants. Figure~\ref{fig:abl_reward} examines the contributions of loss reduction, cellular coverage, and 3D redundancy rewards on cSCC. At a 10\% budget, removing any component lowers PCC for both predictors, while jointly removing the two prior terms further weakens performance. The full reward remains best across the plotted budgets with STRATA; with ST-Net, its advantage is clearest at 10--15\%. These observations are consistent with cellular coverage and redundancy rewards supplying per-spot guidance when round-level loss feedback is sparse, helping balance prototype coverage and spatial diversity under tight budgets. Together, all rewards equip our sampling agent with sparse calibration and prior-driven per-step guidance acting at complementary scales.
\begin{table}[tbhp]
\centering
\scriptsize
\caption{\textbf{Ablation study in STRATA.} Individual branches provide complementary signals, while adaptive fusion consistently improves performance.}
\label{tab:ablation_branches}
\renewcommand{\arraystretch}{1.0}
\setlength{\tabcolsep}{2.5pt}
\resizebox{\columnwidth}{!}{%
\begin{tabular}{l|cc@{\hspace{3pt}}cc}
\toprule
\multirow{2}{*}{\textbf{Model}}
& \multicolumn{2}{c}{\textbf{HER2~\cite{andersson2021spatial}}}
& \multicolumn{2}{c}{\textbf{cSCC~\cite{ji2020multimodal}}} \\
\cmidrule(lr){2-3}
\cmidrule(lr){4-5}
& MSE$\downarrow$ & PCC$\uparrow$
& MSE$\downarrow$ & PCC$\uparrow$ \\
\midrule

\rowcolor{vlmgray}
\multicolumn{5}{c}{\textbf{\textit{Single Branch}}} \\
w.o. 3D $L_{con}$
& \msinline{0.434}{0.132} & \msinline{0.314}{0.053}
& \msinline{1.500}{0.083} & \msinline{0.155}{0.177} \\
STRATA$_{Reg}$
& \msinline{0.404}{0.132} & \msinline{0.328}{0.052}
& \msinline{1.524}{0.080} & \msinline{0.177}{0.194} \\

STRATA$_{Ret}$
& \msinline{0.443}{0.124} & \msinline{0.287}{0.110}
& \msinline{1.431}{0.063} & \msinline{0.190}{0.165} \\

\rowcolor{vlmgray}
\multicolumn{5}{c}{\textbf{\textit{Multiple Branch Fusion}}} \\

STRATA$_{Reg+3D}$
& \msinline{0.386}{0.134} & \msinline{0.338}{0.051}
& \msinline{1.483}{0.090} & \msinline{0.185}{0.189} \\

STRATA$_{Ret+3D}$
& \msinline{0.432}{0.130} & \msinline{0.322}{0.068}
& \msinline{1.348}{0.085} & \msinline{0.219}{0.199} \\

STRATA$_{Reg+Ret}$
& \msinline{0.382}{0.133} & \msinline{0.349}{0.053}
& \msinline{1.375}{0.084} & \msinline{0.216}{0.198} \\

\underline{\textbf{STRATA (Ours)}}
& \cellcolor{resultblue!10}\textbf{\msinline{0.356}{0.123}}
& \cellcolor{resultblue!10}\textbf{\msinline{0.361}{0.049}}
& \cellcolor{resultblue!10}\textbf{\msinline{1.338}{0.078}}
& \cellcolor{resultblue!10}\textbf{\msinline{0.228}{0.153}} \\

\bottomrule
\end{tabular}%
}
\end{table}
\section{Conclusion}
We propose STRATA, a framework that moves beyond conventional 2D fixed-grid sampling and mono-paradigm prediction in spatial transcriptomics, toward budget-aware multi-paradigm adaptive fusion over 3D volumetric structures. STRATA introduces an RL-driven sampling agent on a 3D graph, leveraging single-cell prototype priors to capture cellular rarity and spatial redundancy. Building upon this, a tri-paradigm adaptation network further performs soft routing by assigning preference weights over complementary strategies. Extensive experiments demonstrate that STRATA consistently outperforms current methods. By recasting ST as a co-optimized sampling-prediction problem over 3D voxel volumes, STRATA charts a scalable and cost-effective path toward 3D tissue-level molecular mapping.

\section{Compliance with ethical standards}
\label{sec:ethics}
The private kidney dataset was collected under approval from the Institutional Review Board (IRB) of Vanderbilt University Medical Center (IRB No. 230223).

\section{Acknowledgments}
\label{sec:acknowledgments}
This research was supported by the National Institutes of Health (NIH) under grant R01DK135597, with \$20000, representing 66\% of the total project costs. The content is solely the responsibility of the authors and does not necessarily represent the official views of NIH. This material is based upon work supported by the U.S. National Science Foundation under Award Number 2434229. A total of \$5000 representing 17\% of the total project costs, was financed with this funding. Any opinions, findings, and conclusions or recommendations expressed in this material are those of the author(s) and do not necessarily reflect the views of the U.S. National Science Foundation. A total of \$5000 representing 17\% of the total project costs, was financed with non-governmental funds.


\bibliographystyle{IEEEbib}
\bibliography{strings,refs}

\end{document}